\documentclass[10pt, twocolumn, letterpaper]{article}

\usepackage{cvpr}              % To produce the CAMERA-READY version
\usepackage{graphicx}
\usepackage{amsmath}
\usepackage{amssymb}
\usepackage{booktabs}
\usepackage[accsupp]{axessibility}

\usepackage[pagebackref,breaklinks,colorlinks]{hyperref}

\usepackage{units}
\usepackage[capitalize]{cleveref}
\crefname{section}{Sec.}{Secs.}
\Crefname{section}{Section}{Sections}
\Crefname{table}{Table}{Tables}
\crefname{table}{Tab.}{Tabs.}

\def\cvprPaperID{67} % *** Enter the CVPR Paper ID here
\def\confName{CVPR}
\def\confYear{2023}

\begin{document}

%%%%%%%%% TITLE - PLEASE UPDATE
\title{Spatio-Temporal AU Relational Graph Representation Learning For Facial
Action Units Detection}

% \author{Zihan Wang\\
% College of Computer Science and Software Engineering, Shenzhen University\\
% {\tt\small 2018112013@email.szu.edu.cn}
% \and
% Siyang Song\\
% School of Computing and Mathematical Sciences, University of Leicester\\
% Department of Computer Science and Technology, University of Cambridge\\
% First line of institution address\\
% {ss1535@leicester.ac.uk}
% % For a paper whose authors are all at the same institution,
% % omit the following lines up until the closing ``}''.
% % Additional authors and addresses can be added with ``\and'',
% % just like the second author.
% % To save space, use either the email address or home page, not both
% \and
% Cheng Luo\\
% College of Computer Science and Software Engineering, Shenzhen University\\
% First line of institution2 address\\
% {\tt\small secondauthor@i2.org}
% \and
% Weicheng Xie\\
% First line of institution2 address\\
% {\tt\small secondauthor@i2.org}
% \and
% Linlin Shen\\
% College of Computer Science and Software Engineering, Shenzhen University\\
% First line of institution2 address\\
% {\tt\small secondauthor@i2.org}
% }}

\author{Zihan Wang$^{1,2}$,~
Siyang Song$^{3}$,~
Cheng Luo$^{1,2}$,~
Yuzhi Zhou$^{1,2}$,~
Shiling Wu$^{1,2}$,~\\
Weicheng Xie$^{1,2}$ ~ and ~  
Linlin Shen$^{1,2}$ \thanks{Corresponding Author} \\
\textsuperscript{\rm1}Computer Vision Institute, School of Computer Science $\&$ Software Engineering, Shenzhen University\\
%\textsuperscript{\rm2}Shenzhen Institute of Artificial Intelligence $\&$ Robotics for Society\\
\textsuperscript{\rm2}National Engineering Laboratory for Big Data System Computing Technology, Shenzhen University \\
\textsuperscript{\rm3}School of Computing and Mathematical Sciences, University of Leicester\\
{\tt\small \{2018112013, luocheng2020, 2200271004, 2110276070\}@email.szu.edu.cn,}\\
{\tt\small \{wcxie, llshen\}@szu.edu.cn,}
{\tt\small ss1535@leicester.ac.uk}
}

\maketitle

%%%%%%%%% ABSTRACT
\begin{abstract}
% With the higher requirements for the human affective behavior understanding ability of machine, affective behavior analysis have attracted more and more attention. 

This paper presents our Facial Action Units (AUs) detection submission to the fifth Affective Behavior Analysis in-the-wild Competition (ABAW). Our approach consists of three main modules: (i) a pre-trained facial representation encoder which produce a strong facial representation from each input face image in the input sequence; (ii) an AU-specific feature generator that specifically learns a set of AU features from each facial representation; and (iii) a spatio-temporal graph learning module that constructs a spatio-temporal graph representation. This graph representation describes AUs contained in all frames and predicts the occurrence of each AU based on both the modeled spatial information within the corresponding face and the learned temporal dynamics among frames. The experimental results show that our approach outperformed the baseline and the spatio-temporal graph representation learning allows our model to generate the best results among all ablated systems. Our model ranks at the 4th place in the AU recognition track at the 5th ABAW Competition. Our code is publicly available at \url{https://github.com/wzh125/ABAW-5} .

\end{abstract}

%%%%%%%%% BODY TEXT
\section{Introduction}
\label{sec:intro}

%%% 第一段：研究的定义和背景：介绍一下AU是什么，AU识别的重要性
% In order to achieve artificial intelligence, human affective behavior analysis is an inevitable challenge. Action Units(AU), as a description of facial muscle movements, plays a significant role in affective behavior analysis. AU is defined according to the anatomical characteristics of the face and the muscle movement of the facial expression based on the Facial Action Coding System(FACS)\cite{ekman1978facial}. The different human expression can be deconstructing into the specific multiple Action Units. Facial AU recognition is a multi-label classification task and has received increasing attention as AU is a more objective and comprehensive representation of facial expression.

Human Facial Action Units (AUs), as a widely-used description for facial muscle movements, play a significant role in human behavior understanding \cite{song2020spectral,kollias2019expression,jaiswal2019automatic,song2018human}. Facial AUs are annotated according to the anatomical characteristics of multiple facial muscle movement based on Facial Action Coding System (FACS) \cite{ekman1978facial}. Compared to categorical facial expressions, AUs are more objective and comprehensive representation of facial expressions, and thus drew increasing attentions in recent years \cite{martinez2017automatic}. However, AU detection is a challenging multi-label classification task as AUs are subtle movement of facial muscles, and different facial muscles have different ranges of movement, which are affected by various person-specific factors (e.g., gender and age) as well as contexts (e.g., background, illumination ,and occlusion).
% etc. especially in uncontrolled environments.

%%% 第二段：当前研究方法总结，并提出当前研究方法的不足（这个不足要与本文相关，比如本文要用图，为什么之前不用图的不好）

% The 5th ABAW Competition\cite{kollias2023abaw} provide Action Unit Detection Challenge and the Aff-Wild2 \cite{kollias2019deep,kollias2019expression,kollias2019face,kollias2020analysing,kollias2021affect,kollias2021distribution,kollias2021analysing,kollias2022abaw,zafeiriou2017aff} database will be used. There are several papers in the previous ABAW Competitions, but few work pay attention to the relationships among different AU, they ignore that multiple AU co-occur simultaneously when people generate a certain expression, different AU have underlying relationships that can be represented by graph structure. Also, data imbalance is an issue in Aff-Wild2 database, causing overfitting to AU labels and identities with more samples.

The Action Unit Detection Challenge of the 5th ABAW Competition\cite{kollias2023abaw} is based on the Aff-Wild2 \cite{kollias2019deep,kollias2019expression,kollias2019face,kollias2020analysing,kollias2021affect,kollias2021distribution,kollias2021analysing,kollias2022abaw,zafeiriou2017aff} database. Some of the AU detection approaches in the previous ABAW Competitions \cite{kollias2022abaw,kollias2020analysing,kollias2021analysing} fuse multi-modal features including video and audio to provide multi-dimensional information to predict AUs' occurrence \cite{zhang2022transformer,jin2021multi,jeong2022multi,wang2022action}. Meanwhile, other studies found that AU detection performance can be benefited from multi-task learning\cite{deng2022estimating,jeong2022multi,zhang2021prior,nguyen2023affective}, i.e., jointly conducting expression recognition or valence/arousal estimation provides helpful cues for AU detection. Moreover, temporal models such as GRU \cite{dey2017gate} or Transformer \cite{vaswani2017attention} are also introduced to model temporal dynamics among consecutive frames \cite{nguyen2022ensemble,wang2022action}. While AUs' activation status in each facial display are highly correlated, their relationships provide crucial cues for their occurrence recognition. Meanwhile, the annotations of AUs in the Aff-Wild2 database exhibit a notable imbalance (e.g.,samples of AU7,10,25 are far more than that of AU15,23,24,26 and some AUs only appear on certain identities.), which can result in the training of a biased model that are predisposed to learn AU patterns that have been annotated more frequently in the training set. However, to the best of our knowledge, there is no previous study can jointly address both problems.

% But few work in ABAW Competitions pay attention to the relationships among different AUs, ignoring that multiple AU co-occur simultaneously when people generate a certain expression. Also, data imbalance is an issue in Aff-Wild2 database, causing overfitting to AU labels and identities with more samples.

%%% 第三段：可以提一下IJCAI论文方法的优势可以解决上述问题，但是仍然具有缺陷（比如未考虑时序信息）
Recent studies show that the graph representation is powerful for modelling the underlying relationship among AUs \cite{luo2022learning,song2022gratis,song2021uncertain}. In particular, task-specific multi-dimensional edge features shows strong capability in explicitly describing the relationship between each pair of AUs. 
To overcome the overfitting problem, Ma et al.\cite{ma2022facial}introduces a robust facial representation model MAE-Face for AU analysis. But both of them ignore the temporal information. Considering that the Aff-Wild2 database consists of videos, there is a certain relationship between different frames in a video and the adjacent frames are relatively similar. Therefore, AU detection can be benefited from temporal information.

% Luo et al. \cite{luo2022learning,song2022gratis} proposes an AU relationship modeling approach that deep learns a unique graph to explicitly describe the relationship between each pair of AU of the target facial display. 

\begin{figure*}[ht]
    \centering
    \includegraphics[scale=0.52]{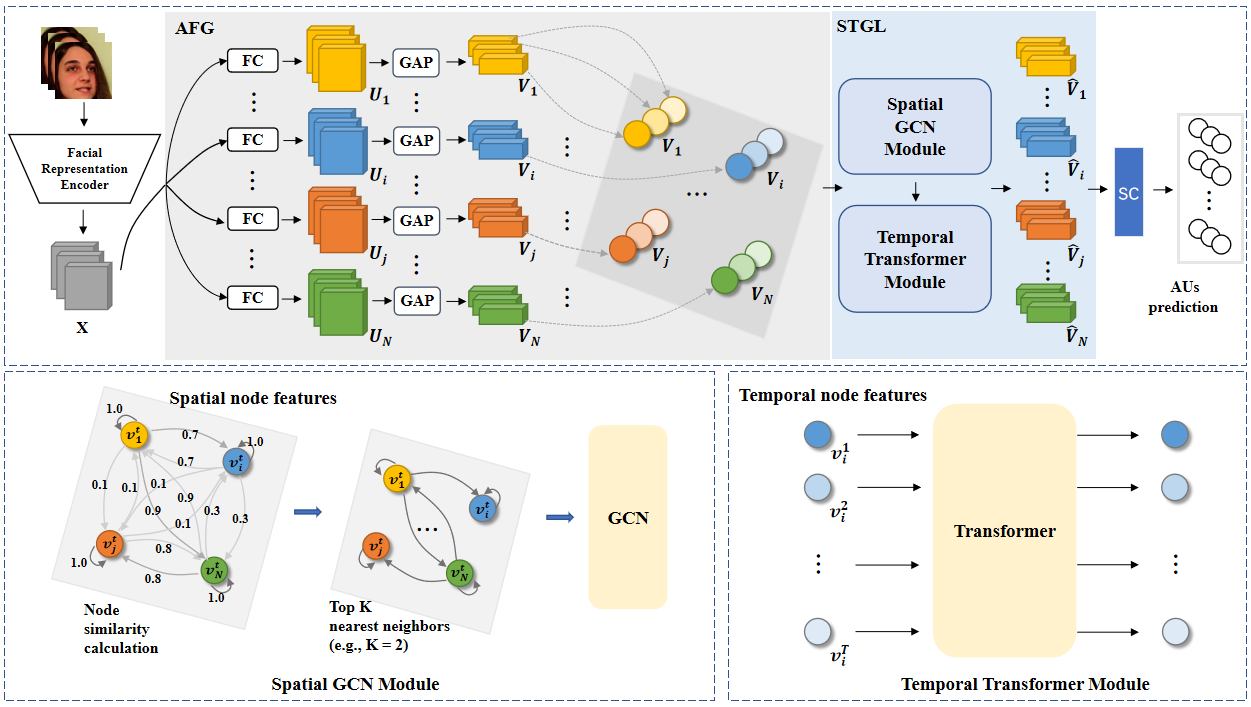}
    \caption{The pipeline of the proposed Spatio-Temporal AU Relational Graph Representation Learning approach}
    \label{fig:overview}
\end{figure*}

%%% 第四段：介绍本文方法的模块，创新性及优势
In this paper, we propose a spatio-temporal facial AU graph representation learning framework for the AU Detection Challenge at the 5th ABAW Competition. Our framework starts with pre-training a masked autoencoder (MAE) \cite{he2022masked} from a set of face databases. This way, the pre-trained MAE can produce a strong facial representation from each input face image. Based on the facial representation, a AU-specific Feature Generator learns specific representation for each AU, which is considered as the node feature in the spatio-temporal AU graph. Then, a spatio-temporal graph learning (STGL) module is introduced to jointly model the spatio-temporal relationships among AUs of all face frames. Specifically, the update of a AU node is not only related to its spatial neighbours in the same frame, but also the nodes in its specific AU sequence, so that the relationships between different AUs and the temporal information of a specific AU sequence can interact and jointly guide the graph to learn representation for each AU node. 
The main contributions of this work are listed as follow:

\begin{itemize}
    \item We pre-train a MAE model based on human face databases, which can generate a strong facial representation from each input facial display, to overcome the data imbalance problem in action units detection.

    \item We propose a spatio-temporal graph learning module to model spacial relationships between different AUs and temporal dependencies among different frames.

    \item The proposed method achieves significant improvement over the baseline and ranks at the 4th place in the AU Detection Challenge at the 5th ABAW Competition.

\end{itemize}

\section{Related Work}

In this section, we systematically review previous AU detection approaches, which are categorized into two types: Non-graph and graph-based AU detection approaches.

%%% 1. 介绍目前的AU 识别方法（非图）

\subsection{Non-graph based AU detection approaches}

Since each AU's activation can only appear in a sparse facial region \cite{kaili2016deep}, several facial region-based methods are proposed \cite{kaili2016deep,li2017eac,li2017action,shao2019facial,jacob2021facial}. For example, Li et al. \cite{li2017eac} first crop key facial regions from face images, and then learn deep features for each facial region individually. Jacob et al. \cite{jacob2021facial} set an attention module to enforce the model to focus on the facial regions corresponding to activated AUs. There are several studies \cite{jaiswal2016deep,chu2017learning,shao2020spatio,nguyen2022ensemble,song2021self} introduce temporal information as the facial muscle movement is a dynamic process. Chu et al. \cite{chu2017learning} use CNN to extract feature from each frame and model the temporal sequence by LSTM. Nguyen et al. \cite{nguyen2022ensemble} utilize Transformer to add temporal information in frame sequence and they ranked 3rd in the AU detection challenge at the ABAW competition 2022. The methods summarised above are built on typical supervised learning, whose generalization capabilities are largely depending on the quality of AU annotations. Subsequently, self-supervised learning strategies \cite{li2019self,ma2022facial,chang2022knowledge,li2019self} recently have been frequently introduced  to AU recognition. In particular, MAE-Face \cite{ma2022facial} first learns a high-capacity model from a large amount of face images without any data annotations, then after being fine-tuned on downstream task including AU detection and AU intensity estimation, which exhibits convincing performance.

% MAE-Face \cite{ma2022facial}that proposed by Ma et al. first learns a high-capacity model from a large amount of face images without any data annotations. Then after being fine-tuned on AU datasets, it exhibits convincing performance for AU recognition. In the 3rd  ABAW competition, Zhang et al. \cite{zhang2022transformer} proposed a multimodal framework to fuse audio and visual feature for AU and expression recognition. Jiang et al. \cite{jiang2022model} pretrain the backbone on different task, including expression and face recognition, and they adopt an ensemble methodology to get the result of AU recognition. Nguyen et al.\cite{nguyen2022ensemble} utilize Gated Recurrent Unit (GRU) and Transformer to introduce temporal information in frame sequence.

\subsection{Graph-based AU recognition approaches}
Considering that relationships between AUs (i.e., AU co-occurrence pattern) play a significant role in AU recognition, some researchers utilize graph neural networks (GNNs) to model the underlying relationship. Li et al.\cite{li2019semantic} is the first attempt that employs the GNN for AU relationships modeling. Song et al. \cite{song2021uncertain} propose an uncertain graph neural network to capture the importance of the dependencies among AUs for each input and estimate the prediction uncertainties. More recently, Luo et al. \cite{luo2022learning,song2022gratis} propose to learn multi-dimensional edge feature-based AU relational graph, where the relationship between each pair of AUs can be explicitly modelled by a task-specific multi-dimensional edge feature. Song et al.\cite{song2022heterogeneous} construct a co-occurrence knowledge graph and a spatio-temporal Transformer module to capture the temporal and spatial relations of AUs. Nguyen et al.\cite{nguyen2023affective} use a facial graph to capture the association among action units for the multi-task learning challenge and they ranked 4th in multi-task challenge at the ABAW competition 2022. These works exhibit the effectiveness of modeling AU relationships.

\section{Methodology}
Given $T$ consecutive facial frames $S = \{f^1, ... ,f^t, ...,  f^T\}$, our goal is to predict AUs' occurrences for each frame. Since there are multiple AUs defined for each facial display, our approach aims to jointly predict multiple AUs for all frame, which is denoted as $P_t = \{p_{1},p_{2},...,p_{N}\} $, where $N$ represents the number of predicted AUs, $t$ denotes the $t_{th}$ frame and $p \in  \{0, 1\}$ can be either activated (1) or inactivated (0). The pipeline of our approach is illustrated in Fig.~\ref{fig:overview}, which consists of a Facial Representation Encoder (FRE) described in Sec.~\ref{subsec:FRE}, an AU-specific Feature Generator (AFG) described in Sec.~\ref{subsec:AFG} and a Spatio-Temporal Graph Learning (STGL) module described in Sec.~\ref{subsec:STGL}. Specifically, the FRE first extracts global facial features from each image of the input face sequence. Then, the AFG individually learns a representation for each AU of each facial frame based on its global representation. This way, a spatio-temporal facial graph representation can be constructed. Finally, the STGL module takes the spatio-temporal facial graph as the input and jointly predicts all AUs' occurrence for all frames, where both spatial facial displays and temporal facial dynamics are considered.

\subsection{Facial Representation Encoder}
\label{subsec:FRE}
Masked autoencoder (MAE) \cite{he2022masked} is a self-supervised learned model which reconstructs original images from a set of masked images. It is made up of a linear projection layer, a 12-layer encoder and a 4-layer decoder that were defined by the Vision Transformer \cite{dosovitskiy2020image}. The well-trained  MAE can be fine-tuned for various downstream tasks. 

\begin{figure}
    \centering
    \includegraphics[scale=0.5]{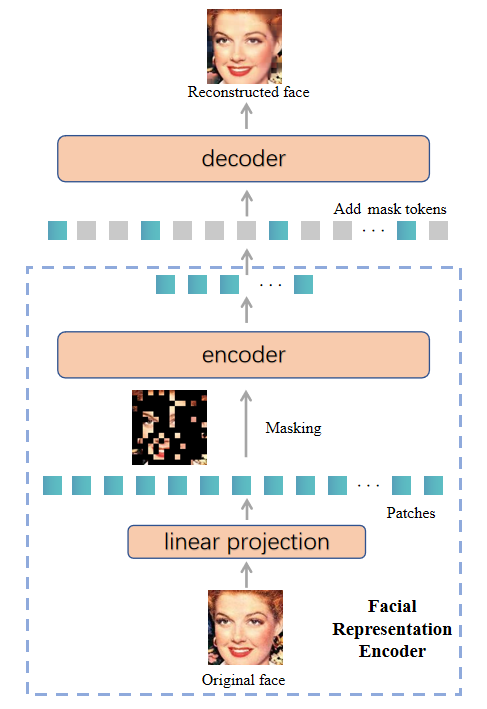}
    \caption{ Illustration of the masked autoencoder model}
    \label{fig:mae}
\end{figure}

Since MAE has a strong representation learning capability and scalability, we propose to first pre-train a MAE model using a large amount of face images from CASIA-WebFace\cite{yi2014learning}, AffectNet\cite{mollahosseini2017affectnet}, IMDB-WIKI \cite{rothe2018deep} and CelebA\cite{liu2015deep}, making the pre-trained MAE to be able to generate strong facial representations from previous unseen face images. This pre-training strategy would not only help the model to alleviate the data imbalance problem in target AU database, but also improve the generalization ability of network in uncontrolled environments. Fig. \ref{fig:mae} illustrates the MAE model's pre-training, where randomly masked face images are fed to the encoder to generate latent features, and then the decoder reconstructs the original image from these latent features. 

Subsequently, the linear projection layer and the encoder of the pre-trained MAE can generate a strong representation for each input face image, which are employed as the encoder for our facial AU recognition pipeline. To generate AU predictions for the input facial image sequence $S = \{f^1, ... ,f^t, ...,  f^T \} \in \mathbb{R}^{T\times C\times H\times W}$ , the linear projection layer first encodes each frame to a set of patches, which are treated as a set of tokens to be fed into the pre-trained transformer encoder without masking operation. As a result, a set of facial representations $X = \{ x^1,...,x^t,..., x^T \}\in \mathbb{R}^{T\times m \times d}$ can be generated, where each $x_t \in X$ represents a global facial representation of a face image; $m$ is the number of patches and $d$ denotes the dimension of each patch.

\subsection{AU-specific Feature Generator}
\label{subsec:AFG}
Since each AU's activation only appears in a specific local facial region but will be reflected bu other facial regions, we propose a AU-specific Feature Generator(AFG) to extract unique feature for each AU from the global facial representation $X$. In particular, the AFG consists of $N$ branches, where each is made up of a fully connected layers (FC) followed by a global average pooling (GAP) layer. The $i_{th}$ FC layer of first projects the $X$ to an AU-specific feature map $U_i \in \mathbb{R}^{T \times m \times d} $, and then GAP layer yields $T$ vectors consisting  $V_i = \{ v_{i}^{1},...,v_{i}^{t},...,v_{i}^{T} \} \in \mathbb{R}^{T \times d}$, where $v_{i}^{t}$ denotes the representation of the $i_{th}$ AU in the $t_{th}$ frame.

\subsection{Spatial-Temporal Graph Learning}
\label{subsec:STGL}
As discussed before, AUs in each facial display are related to each other. Meanwhile, since human facial behaviours are continuous and smooth, AU activation status in adjacent frames are also temporally correlated. In this sense, our method jointly learns both the spatial relationship among AUs within each face frame as well as the their temporal relationship among face frames. Specifically, the Spatial-Temporal Graph Learning (STGL) module consists of a spacial GCN module for spatial AU relationship modelling and a temporal transformer module for temporal AU relationship modelling.

\subsubsection{Spacial GCN module}

Firstly, we employ the Facial Graph Generator (FGG) proposed by \cite{luo2022learning,song2022gratis} to learn a spatial AU graph representation for each face frame, which consists of $N$ nodes describing features of the $N$ target AUs. Then, the connectivity (edge presence) between each pair of nodes is defined according to the similarity of their features, i.e., each node connects with its $K$ nearest neighbour nodes with highest similarity scores. This way, the topology of the generated graph representation would have adapted topology for different facial displays. After that, a GCN layer is adopted to update node features for the obtained facial graph representation. The new representation of the $i_{th}$ AU in the $t_{th}$ frame can be calculated by its neighbours in the spacial dimension as: 
\begin{equation}
    v_{i}^{t} =  \sigma [v_{i}^{t} + g(v_{i}^{t} ,\sum_{j=1}^{N}r(v_{j}^{t},e_{i,j}^{t}) )]
\end{equation}
where $\sigma$ is the activation function; g and r denotes differentiable functions of the GCN layer, and $e_{i,j}^{t} \in \{ 0 ,1 \}$denotes the connectivity between $v_{i}^{t}$ and $v_{j}^{t}$. Specifically, the above operations will be conducting in each frame of the input sequence.

\subsubsection{Temporal transformer module}

Since transformer \cite{vaswani2017attention} is a superior model to learn long and short-range temporal dependencies, we then propose to utilize the transformer to update AU representations by considering temporal dynamics among facial frames. In temporal dimension, the graph nodes output from the GCN layer are considered as $N$ sequences for $N$ AUs, each of which consists of $T$ nodes. For the $i_{th}$ AU sequence $V_i = \{ v_{i}^{1},\cdots,v_{i}^{t},\cdots,v_{i}^{T} \}\in \mathbb{R}^{T \times d}$, the $T$ nodes will be taken as $T$ tokens which are then fed to the transformer. In particular, each AU node sequence $V_i$ is then individually updated as $\hat{V}_i$ as follows:
\begin{equation}
\begin{split}
    & \hat{V}_i = Z + \text{FFN}(\text{LayerNorm}(Z)) \\  
    & Z = V_i + \text{Att}(Q,K,V)  \\
    & Q = V_i W_Q, ~~ K = V_i W_K,  ~~ V = V_i W_V   
\end{split}
\end{equation}
where FFN is the feed forward network in transformer; Att denotes the self-attention function; and $W_Q$, $W_K$ and $W_V$ are trainable weight matrices. We perform these operations on each AU sequence in the temporal dimension.

In this paper, three Spatio-Temporal Graph Learning (STGL) modules are stacked to produce spatio-temporal AU graph representations. Then, the similarity calculating (SC) strategy \cite{luo2022learning} is employed to predict the probability. For the $i_{th}$ AU in the $t_{th}$ frame, a trainable vector $s_i$ which has the same dimension as $\hat{v}_{i}^{t}$ is shared across all frames, and the prediction can be denoted as:
\begin{equation}
 P_{i}^{t} = \frac{\sigma(\hat{v}_{i}^{t})^T \sigma(s_i)}{|| \sigma(\hat{v}_{i}^{t}) ||_2 || \sigma(s_i)||_2}   
\end{equation}
where $\sigma$ is the activation function.

\begin{table*}[h]
    \centering
    \begin{tabular}{cccccccccccccc}
    \toprule
        &AU1&AU2&AU4&AU6&AU7&AU10&AU12&AU15&AU23&AU24&AU25&AU26&Average  \\
    \midrule
    Baseline\cite{kollias2023abaw}
    & -& -& -& -& -& -& -& -& -& -& -& -& 39.0 \\
    ME-Graph\cite{luo2022learning} &
    56.1&44.9&52.1&61.7&74.8&\textbf{75.4}&73.4&28.4&18.4&12.4&84.5&32.2&51.2 \\
    Netease\cite{zhang2023facial} &
    55.3&\textbf{51.4}&\textbf{56.7}&\textbf{67.3}&\textbf{75.8}&75.1&\textbf{75.8}&31.2&17.3&\textbf{33.8}&83.9&\textbf{42.3}&\textbf{55.9} \\
    Ours & 
    \textbf{57.8}&48.0&55.9&61.9&75.5&74.6&72.0&\textbf{35.6}&\textbf{21.6}&23.7&\textbf{86.0}&38.3&54.3 \\
    
    \bottomrule
    \end{tabular}
    \caption{F1 score (in \%) results achieved for 12 AUs on validation set.The
highest scores are indicated in bold.}
    \label{tab:result}
\end{table*}

\begin{table*}[h]
    \centering
     \resizebox{1\textwidth}{!}{
    \begin{tabular}{c|ccc|cc|c}
    \toprule
             & Swin-B (ImageNet)    &MAE (ImageNet) & MAE (hybrid dataset)  & Spatial    &  Temporal  &  Average F1-score \\

    \midrule
    (i) &\checkmark      &                                     &                             &                & &49.4\\
    (ii) &\checkmark      &                                     &                             &  \checkmark     & &50.7\\
    (iii) &\checkmark      &                                     &                             &  \checkmark     &  \checkmark &52.6\\
  (iv) &   & \checkmark    &                  &      &     & 48.2\\
  (v) &   & \checkmark    &                  &\checkmark      &     & 49.5\\
    
    (vi) &   & \checkmark    &                  &\checkmark      &  \checkmark     &50.0\\

    (vii) &   &                    &  \checkmark                 &                              &   & 52.4  \\
     (viii) &   &                    &  \checkmark                 &   \checkmark                            &   & 52.7  \\
    (ix)&   &                    & \checkmark                  &\checkmark  &\checkmark               &  54.3\\    
    \bottomrule
    \end{tabular}
    }
    \caption{Ablation study on validation set.}
    % \caption{The F1 score (in \%) of ablation study on validation set.}
    \label{tab:my_label}
\end{table*}

\subsection{Loss Function}

We propose a two-stage training strategy to train our AU detection model. At the first stage, we first pre-train the transformer-based facial representation encoder by reconstructing the masked patches of the original face images. Here, we employ Mean Square Error (MSE) loss to constrain the difference between the reconstructed patches and the original patches at the pixel-level. Suppose that $M$ patches are masked at the beginning, the pre-training loss $L_\text{pre}$ is formulated as:
\begin{equation}
L_\text{pre} = \sum_{m=1}^{M}(\hat{p}_m - p_m)^2 \tag{4}
\end{equation}
where $\hat{p}_m$ denotes the ground truth pixels and $p_m$ denotes the reconstructed pixels.

Since AU detection is a multi-label binary classification problem, and most AUs are inactivated for the marjority of face frames, we use an asymmetric loss to optimize the network at the second training stage, which enforce the whole framework to jointly output multiple AUs occurrence prediction. The $L_{au}$ is denoted as:
\begin{equation}
L_{au} = - \sum_{i=1}^{N} \sum_{t=1}^{T} [y_{i}^{t} \log(p_{i}^{t}) + p_{i}^{t} (1-y_{i}^{t})  \log(1-p_{i}^{t}) ]\tag{5}
\end{equation}
where $p_{i}^{t}, y_{i}^{t}$ are the prediction and ground truth; $N$ and $T$ are the numbers of AUs and frames of the input face sequence, respectively. The first $p_{i}^{t}$ in $p_{i}^{t} (1-y_{i}^{t})  \log(1-p_{i}^{t})$ can be considered as the weight of negative samples(inactivated AUs), which down the loss values caused by inactivated AUs that are easy to detect,enforcing the training process to focus on activated AUs and
inactivated AUs that are hard to be correctly recognized.

\section{Experimental results}

% Our training process can be divided into two stages. At the first stage, we leverage the state-of-the-art self-supervised strategy, MAE \cite{he2022masked} to pre-train our network on a hybrid face dataset, including CASIA-WebFace\cite{yi2014learning}, AffectNet\cite{mollahosseini2017affectnet}, IMDB-WIKI \cite{rothe2018deep} and CelebA\cite{liu2015deep}. 
% It allows the network to be initialized with a good priori of facial features.
% After that, the network is trained on the ABAW dataset to be capable of detecting action units.

% In this section, we will introduce the dataset details, experimental settings and our results on the validation set. To prove the effectiveness of each module, we also provide ablation studies.

\subsection{Dataset}

\textbf{Dataset for MAE:} Self-supervised pre-training helps neural networks learn effectively discriminative representations, however, it will bring limited gains for downstream tasks if the size of training data is limited. 
For this reason, we collect a hybrid face dataset from different in-the-wild datasets, following the collecting method in the prior work \cite{ma2022facial}.
This hybrid dataset involves four subsets from  
CASIA-WebFace\cite{yi2014learning}, AffectNet\cite{mollahosseini2017affectnet}, IMDB-WIKI \cite{rothe2018deep} and CelebA\cite{liu2015deep}, respectively.
These source datasets are well-known and widely used in fields spanning from face recognition to expression recognition.
However, we found some low-quality image data included in these datasets.
In order to make the training more effective, we remove all images with blur and incomplete faces.
Finally, we can obtain the hybrid dataset of around 1,920,000 face images without annotations for our pre-training.
% are combined to form the pre-training dataset.

% A large amount of image data are necessary for self-supervised pre-training. So We follow the approach of \cite{ma2022facial} to collect a hybrid face dataset from different in-the-wild datasets. The hybrid set consists of CASIA-WebFace\cite{yi2014learning}, AffectNet\cite{mollahosseini2017affectnet}, IMDB-WIKI \cite{rothe2018deep} and CelebA\cite{liu2015deep}, which are commonly used for face recognition or expression recognition task. All blur and incomplete faces are removed and totally about 1,920,000 face images without annotations are combined to form the pre-training dataset.

\textbf{Dataset for AU detection:} The AU Detection Challenge at 5th ABAW Competition \cite{kollias2023abaw} provides 541 video sequences from Aff-Wild2 dataset.
Each frame of a video sequence in this dataset is manually or automatically annotated with labels of 12 AUs, namely AU1, AU2, AU4, AU6, AU7, AU10, AU12, AU15, AU23, AU24, AU25, and AU26.
Totally, this dataset contains 2,627,632 frames,  with 438 subjects, 268 of which are males and 170 are females. Meanwhile, this video dataset is split into a training set of 295 sequences, a validation set of 105 sequences, and a testing set of 141 sequences in a subject-independent manner.

% Besides, the videos are split into training, validation and testing sets in a subject independent manner.

% that contains annotations in terms of 12 AUs, namely AU1, AU2, AU4, AU6, AU7, AU10, AU12, AU15, AU23, AU24, AU25 and AU26. In total, 2, 627, 632 frames, with 438 subjects, 268 of which are male and 170 female, have been annotated manually and automatically. Besides, the videos are split into training, validation and testing sets in a subject independent manner.

\subsection{Experimental settings}

\textbf{Details for MAE pre-training:} We first leverage RetinaFace \cite{deng2020retinaface} to perform face detection and alignment for each image from the hybrid dataset and crop it to $256 \times 256$. The encoder and decoder  are initialized with the weights pre-trained on ImageNet-1k \cite{deng2009imagenet} dataset. 
When reconstructing each masked image, we applied random cropping augmentation and chose a mask ratio of $75\%$.
During training, a AdamW optimizer with the learning rate of $1.5e^{-4}$ is used, with batch size of 512, and weight decay of $0.05$.
Totally, we pre-train our network for 300 epochs, 40 of which are warm-up epochs.

% During reconstructing, Random cropping and a mask ratio of $75\%$ are applied, with the learning rate of $1.5e-4$ and a weight decay of $0.05$. We pre-train MAE for 300 epochs, with 40 warmup epochs. The batch size is set to 512 and the AdamW optimizer is also used.

\textbf{Details for AU detection training:} 
At this training stage, we follow the cropped-and-aligned version of the Aff-Wild2 dataset. Each subject from the training set is recorded with one video sequence. During training, we randomly select a video clip of 16 frames as input to our model.
For validating and testing, we split each video data into segments, each of which contains 16 frames.
If the number of frames of the segment is less than 16, we supplement it with blank frames.
During the training process, we employ an AdamW optimizer with a weight decay of $5e^{-4}$. The number $K$ for choosing the nearest neighbors is set to 4. The learning rate is set to $1e^{-4}$ and adjusted by a cosine decay learning rate scheduler.

% In training set, each subject have a video, in which we randomly select 16 continuous frames as a training sample in one epoch. In validation and test set, all videos are split to segments with 16 frames, with the vacant positions filled with the previous frame. During training, we employ an AdamW optimizer with a weight decay of $5e^{-4}$. The number K for choosing nearest neighbors is 4. The learning rate is set to 1e-4 and scheduled with cosine decay. We train the model for 10 epochs and the batch size is set to 16.

\textbf{Evaluation metrics:}
We evaluate the AU detection performance of methods by the average F1-score across all AUs. This metric is defined as: 
\begin{equation}
F_1^{AU} = \cfrac{\sum_{i=1}^{N} F_1^{AU, i}}{N}\tag{6}
\end{equation}
where N denotes the number of AUs, and F1-score $F_1^{AU, i}$for individual AU class is computed as:
\begin{equation}
    F_1^{AU,i} = 2 \cdot  \frac{ P^{AU,i} \cdot  R^{AU,i} }{ P^{AU,i} +  R^{AU,i} } \tag{7}
\end{equation}
where $P^{AU,i}$ is the calculated precision for the $\text{i}_{th}$ AU and  $R^{AU,i}$ is the recall rate for it.

% $$F_1^{AU,n} = 2 \times \frac{ { P } \cdot  {Recall}}{ {Precision}+  {Recall.}}$$

% $$F_1^{AU,n} = 2 \times \frac{ { Precision } \cdot  {Recall}}{ {Precision}+  {Recall }}$$

% For AU detection, the performance is measured using the average F1 Score across all AUs. Therefore, the evaluation criterion is defined as :
% $$F_1^{AU} = \frac{\sum_{n=1}^{N} F_1^n}{N}$$

\subsection{Results on validation set}
% The $5_{th}$ ABAW competition provides the official training and validation set. We split each validation video as several segments with fixed length and make a prediction for each frame in a segment. 

Tab.~\ref{tab:result} presents the evaluation results of AU detection on the validation set, reporting the F1-score for each AU. Our method achieves notable improvement over the baseline, increasing the average F1-score from 39.0 to 54.3. Moreover, we compare our results with those of ME-Graph\cite{luo2022learning}, and our method outperforms theirs by an average F1-score of 3.1. While our overall result is slightly lower than that of the first-place team Netease, we achieved higher F1-score than them in some AU categories. These results demonstrate the effectiveness of our approach in detecting AUs.

\subsection{Results on test set}
The final results of Action Unit Detection Challenge on test set are presented in the Tab.~\ref{tab:result on test set}. Specifically,we achieved an average F1-score of $51.3$, which places us in 4th position with only a slight difference from the third-place team's score of 51.4. Netease Fuxi Virtual Human and SituTech, who took first and second place respectively, used a similar approach to ours by employing a pre-trained model to extract facial features.  Additionally, Netease Fuxi Virtual Human leveraged the multi-modal and temporal information from the videos and implemented a transformer-based framework to fuse the multi-modal features. SituTech also incorporated audio information and employed several ensemble strategies. Meanwhile, the third-place team focused on extracting facial local region features related to AU detection and also utilized a graph neural network to model the relationship between AUs.

\begin{table}
    \centering
    \begin{tabular}{c|c}
    \toprule
       Teams  &  F1-score(in \%)\\
       \midrule
       Netease Fuxi Virtual Human \cite{zhang2023facial} & 55.5\\
       SituTech & 54.2\\
       USTC-IAT-united \cite{yu2023local}& 51.4\\
       \textbf{SZFaceU (Ours)} & \textbf{51.3}\\
       PRL  \cite{vu2023vision}& 51.0\\
       CtyunAI \cite{zhou2023continuous}& 48.9\\
       HSE-NN-SberAI \cite{savchenko2023emotieffnet}& 48.8\\
       USTC-AC \cite{wang2023facial}& 48.1\\
       HFUT-MAC\cite{zhang2023facial1} & 47.5\\
       SCLAB-CNU \cite{nguyen2023transformer}& 45.6 \\
       USC-IHP\cite{yin2023multi} & 42.9\\
       Baseline\cite{kollias2023abaw} & 36.5\\
       \bottomrule
    \end{tabular}
    \caption{Action Unit Detection Challenge Results on test set}
    \label{tab:result on test set}
\end{table}

\subsection{Ablation study}
Tab.~\ref{tab:my_label} presents the results of our ablation studies.We choose swin transformer\cite{liu2021swin} as our baseline.
We can observe that the proposed method of incorporating spatial AU graph learning, as shown in (ii), (v), and (viii), leads to significant improvements over the models that lack this feature, as indicated in (i), (iv), and (vii), respectively. Similarly, the inclusion of temporal graph learning, as demonstrated in (iii), (vi), and (ix), yields substantial gains over the models without it, as demonstrated in (ii), (v), and (viii), respectively. These findings highlight the crucial role of modeling the temporal relationships between successive facial frames and the spatial relationships among different AUs in enhancing the accuracy of AU detection.
Furthermore, the model pre-trained on ImageNet using vanilla MAE (vi) is not capable of performing better than the baseline (iii). The possible reason could be that there is a  significant domain gap between the dataset for universal object recognition and the dataset for facial tasks. 
However, when we replace the MAE pre-train dataset with the collected hybrid dataset, the model (ix) shows superior performance ($54.3$ average F1-score) than the baseline (iii) ($52.6$ average F1-score).

% To prove the effectiveness of MAE, we choose swin transformer as our backbone and make comparisons. 

% Swin transformer are pre-trained on ImageNet\cite{deng2009imagenet} and MAE in our hybrid face dataset, and then both add the spacial graph module, the AU average F1 score of two model are $50.7\%$ and $51.6\%$. After adding the temporal module, the AU average F1 score of them are $52.6\%$ and $54.3\%$，respectively. But when MAE is pretrained on ImageNet, it present a relatively bad performance. Therefore, the MAE with self-supervised learning is more advantageous. Then we consider the improvement that the temporal module brings. The data show that the average F1 score increases from $50.7\%$ to $52.6\% $when using MAE, and from $51.6\%$ to $54.3\%$ when backbone is swin transformer after introducing temporal module, which means that temporal information is quite beneficial for AU detection.

\section{Conclusion}
This paper proposes an effective spatio-temporal AU relational graph representation learning method for AU occurrence recognition, where MAE is introduced as the facial representation encoder. Experimental results demonstrate that the proposed approach achieved excellent performance in jointly detecting multiple AUs in face videos, which ranked at the 4th place at the 5th Affective Behavior Analysis in-the-wild (ABAW) Competition.

%%%%%%%%% REFERENCES
{\small
\bibliographystyle{ieee_fullname}
\bibliography{egbib}
}

\end{document}